# Graph Neural Network Enhanced Sequential Recommendation Method for Cross-Platform Ad Campaigns

*Xiang Li[1,*], Xinyu Wang[2,a], Yifan Lin[3,b]*

*[1] Department of Electrical and Computer Engineering, Rutgers University, Piscataway, NJ, USA 08854*

*[2] School of Global Public Health, New York University, New York, NY, USA 10003*

*[3] Pratt School of Engineering, Duke University, Durham, NC, USA 27708*

*[*] xl470@scarletmail.rutgers.edu*

*[a] xw3080@nyu.edu*

*[b] yifan.lin@alumni.duke.edu*

*Abstract*— **In order to improve the accuracy of cross-platform advertisement recommendation, a graph neural network (GNN)-based advertisement recommendation method is analyzed. Through multi-dimensional modeling, user behavior data (e.g., click frequency, active duration) reveal temporal patterns of interest evolution, ad content (e.g., type, tag, duration) influences semantic preferences, and platform features (e.g., device type, usage context) shape the environment where interest transitions occur. These factors jointly enable the GNN to capture the latent pathways of user interest migration across platforms. The experimental results are based on the datasets of three platforms, and Platform B reaches 0.937 in AUC value, which is the best performance. Platform A and Platform C showed a slight decrease in precision and recall with uneven distribution of ad labels. By adjusting the hyperparameters such as learning rate, batch size and embedding dimension, the adaptability and robustness of the model in heterogeneous data are further improved.**

## Introduction

Graph Neural Network (GNN) as a deep learning method that can effectively process graph-structured data, has been widely used in the fields of complex networks, recommender systems and social media analysis. Its ability to capture deep dependencies between data by modeling relationships between nodes and edges makes it possible to accurately capture various types of data features and potential correlations when dealing with large-scale heterogeneous data. In the field of advertisement recommendation, GNN has a significant advantage in improving the effectiveness and accuracy of recommendation systems by mining the multidimensional relationships among user behavior, advertisement content and platform features. Based on this, exploring the application of graph neural networks in cross-platform advertisement recommendation has important theoretical and practical value, as it deepens the understanding of interest migration modeling in heterogeneous graph structures (theoretical), and enhances the precision and adaptability of ad delivery strategies across diverse platforms (practical).

## Characterization of Cross-Platform Advertising Campaigns

Cross-platform advertising campaigns involve the integration and analysis of data from multiple platforms and need to pay attention to the similarities and differences in user behavior across platforms[1]. The data characteristics of each platform include users' click rate, interaction frequency, browsing time, and device usage habits, all of which contribute differently to ad effectiveness and recommendation accuracy: higher click rates and interaction frequency often indicate stronger engagement and are directly linked to higher conversion potential, while longer browsing time reflects sustained attention to ad content. Device usage habits offer context about access environments, influencing ad display strategies and user receptiveness, which ultimately shape the effectiveness and acceptance of ad campaigns. Through graph neural network (GNN), deep learning and feature extraction can be performed on cross-platform user behavior data to capture the correlation between platforms and improve the accuracy of the recommendation system. Effective campaign analysis requires consideration of the diversity of advertising content, user profiles and platform characteristics to optimize the advertising strategy and further enhance the prediction ability of advertising effects.

## Cross-Platform Ad Campaign Recommendation Model Construction Based on Graph Neural Network

### Graph Neural Network Model Architecture Design

In constructing a graph neural network architecture for cross-platform advertisement recommendation, the heterogeneity and dynamics of the user-platform-advertisement ternary relationship must be fully considered. The model design employs a combined architecture of multilayer graph convolutional network (GCN) and graph attention network (GAT) to support non-uniformly weighted learning of node features and dynamic representation updating of neighboring edges. Specifically, GCN captures global structural information by aggregating features of neighboring nodes, which is crucial for understanding relationships in large-scale graph structures. GAT, on the other hand, introduces an attention mechanism that enables the model to assign different weights to neighboring nodes, thus enabling more fine-grained and adaptive learning of relationships, especially in cases where the graph structure is sparse or highly heterogeneous (e.g., cross-platform advertisement data). In this study, we use a hybrid model of Graph Convolutional Network (GCN) and Graph Attention Network (GAT), where GCN is used for feature aggregation of the global structure, while GAT weights the contribution of each neighboring node through the attention mechanism, which in turn improves the accuracy of cross-platform advertisement recommendation. Mathematical expression of feature aggregation strategy:

For each layer of Graph Convolutional Network (GCN) and Graph Attention Network (GAT) model, the mathematical expression of feature aggregation is as follows:

$$h_{\upsilon}^{(l+1)} = \sigma\left(\sum_{u \in N(\upsilon)} \frac{1}{c_{\upsilon u}} W^{(l)} h_u^{(l)}\right) \quad (1)$$

Where $h_{\upsilon}^{(l+1)}$ denotes the feature representation of node $\upsilon$ in layer $l+1$, $N(\upsilon)$ is the set of neighboring nodes of node $\upsilon$, $c_{\upsilon u}$ is the normalization coefficient between node $\upsilon$ and node $u$, $W^{(l)}$ is the weight matrix in layer $l$, and $\sigma$ is the activation function. In GAT, $1/c_{\upsilon u}$ is replaced by the weights computed through the attention mechanism, thus enabling the model to dynamically adjust the importance of each neighbor node[2] .The input graph is modeled as a heterogeneous graph, and the node types cover three categories: users, advertisement spots, and platform devices, among which the user nodes contain behavioral features in 11 dimensions such as Click Frequency, Active Duration, and Historical Interaction Count; and the advertisement nodes contains 8 dimensions of content characteristics such as placement type, content label, material duration, etc.; the platform node is coded with 6 dimensions such as device type, system version, average daily visit duration, etc.**Error! Reference source not found.** . Meanwhile, in order to better capture the evolution of user interests over time, a time-aware GNN (e.g., TGAT) is introduced into the model, which enables the model to adapt to the dynamically changing interest migration over time by encoding user behaviors into time windows and modeling the sequence. Specifically, user behavior data is partitioned into multiple time windows (e.g., Δt = 2h, 6h, 12h) and temporal information is incorporated into node features through temporal encoding. In addition, the time-aware graph attention network (TGAT) is able to adaptively adjust the weights of user behaviors within the time windows to strengthen the model's ability to capture changes in user interests,As shown in Figure 1.

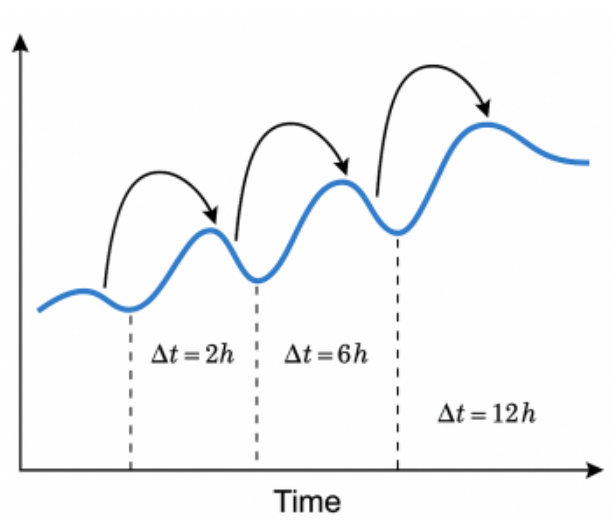

Figure 1. Temporal User Behavior with Time Windows (Δt = 2h, 6h, 12h)

Furthermore,To better model cross-platform user interest transitions, we adopt a shared user node strategy, where the same user across different platforms is represented as a single node. User behavior logs from different platforms are aligned via a unified user ID mapping mechanism based on hashed identifiers and temporal activity matching. In the graph edge design, we introduce a new edge type "view-cross-platform", connecting the same user node across platform nodes when the user interacts with semantically similar ads within a predefined time window (e.g., 24 hours). The edge weight reflects the frequency and continuity of cross-platform behaviors (e.g., 0.85 if the user clicked similar ads within 24h across platforms). This shared structure enhances the GNN's ability to capture interest migration paths and semantic consistency across heterogeneous platforms.

The graph edge types cover three categories: "view-platform", "click-ad", and "browse-user", and the edge attributes cover three categories: "watch-platform", "click-ad", and "browse-user". The edge attributes include 12 numerical attributes such as timestamp, click-conversion rate grouping, ad weight index, etc. These attributes are first normalized to ensure that they are on a comparable scale before being input into the model. Specifically, numerical attributes like timestamps are standardized to a range between 0 and 1 using min-max normalization, while click-conversion rates and ad weight indices are scaled using z-score normalization to center the data and reduce variance. This preprocessing step ensures that the model can efficiently learn from the data without being biased by any single attribute's scale. As shown in Fig. 1, the graph edge type covers "view-platform", "click-ad", and "view-user". As shown in Fig. 2.

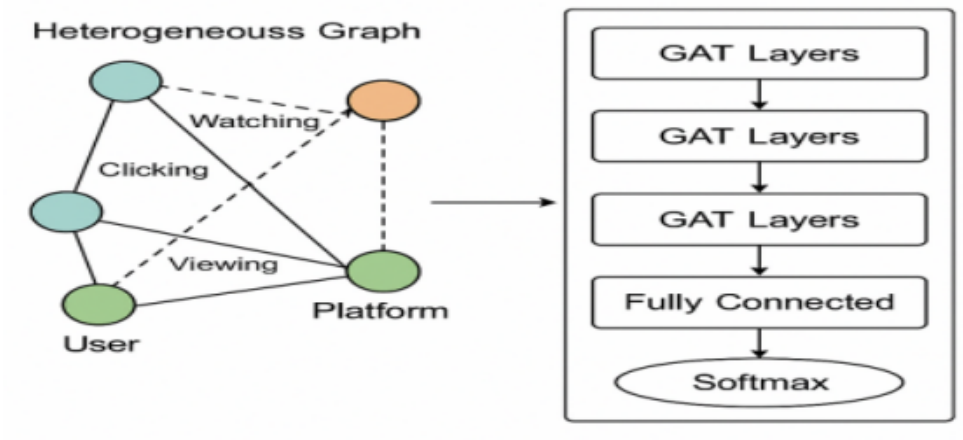

Figure 2. GNN-based cross-platform ad recommendation architecture

## Hyperparameter Optimization for Sequential Recommendation Models

To optimize cross-platform ad recommendation performance, hyperparameters are finely tuned, including learning rate (0.0005, 0.001, 0.0050), batch size (128, 256, 384, 512), embedding dimensions (64, 128, 256), and number of graph attention heads (GAT Heads). A combination of grid search and Bayesian optimization is used to adjust the learning rate and fix the minimum validation set loss point. This approach is advantageous because grid search allows for an exhaustive exploration of hyperparameters over a fixed, predefined grid, ensuring that a broad range of potential values is considered. On the other hand, Bayesian optimization intelligently narrows down the search space by using previous evaluation results to predict the next best hyperparameter settings, thus improving efficiency and reducing computational cost[3] . The batch size is adjusted according to the platform user active density, and the embedded dimension test avoids GNN gradient explosion. Optimization is carried out under different ad type share and platform behavior time window (Δt=2h, 6h, 12h) to improve recommendation performance and stability, and the sequence recommendation score is calculated as follows:

$$S_{i,j} = \sum_{t=1}^{T} \alpha_t \cdot f_\theta(h_t, e_{t,j}) \qquad (2)$$

Among them, $S_{i,j}$ denotes the recommendation score of the user $i$ for the ad $j$ , $T$ is the length of the user behavior sequence, $\alpha_t$ is the GAT attention weight, $h_t$ is the user state vector of the $t$ th moment, $e_{t,j}$ is the feature embedding of the ad at that moment, and $f_\theta(\cdot)$ is the feed-forward network scoring function. By dynamically adjusting the $\alpha_t$ mechanism, the model can adapt to different user behavior lengths and improve the accuracy of cross-platform behavior transfer capture[5] . The effect of related parameter configuration is shown in Table 1.The complete optimization and inference process is visualized in Fig. 3, which outlines the modular pipeline including data input, model configuration, batch parallelism, and multi-hop subgraph inference.

TABLE I. HYPERPARAMETER COMBINATION DESIGN AND PERFORMANCE PARAMETERS OF THE SEQUENCE RECOMMENDATION MODEL

| parameter term | parameter value | Scope of adjustment | Sample of data inputs (10,000) | GPU Occupancy (MB) | Model training time (min/epoch) |
|---|---|---|---|---|---|
| learning rate | 0.0005/0.001/0.005 | three-speed adjustment | 42/42/42 | 7632 | 17/21/25 |
| Batch size | 128/256/384/512 | four-speed adjustment | 30/35/42/48 | 6520~9100 | 14/16/18/21 |
| Embedding Dimension | 64/128/256 | three-speed adjustment | 42 | 5980~8124 | 15/17/20 |
| attention span | 4/8/12 | Multiple Attention Optimization | 42 | 6220~8910 | 16/19/23 |

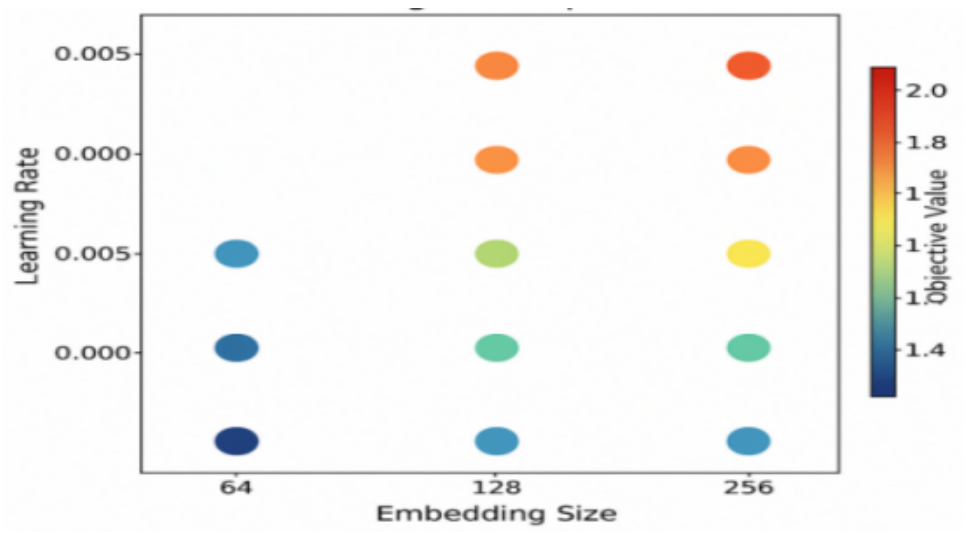

Figure 3. Learning rate and embedding size optimization parameter grid

## Model Training and Inference Strategies

The model training phase adopts a distributed training strategy, which is accelerated by four NVIDIA A100 40GB graphics cards in parallel to realize the convergence regulation of the ultra-large scale heterogeneous graph structure in the high-dimensional feature space[6] . The input data dimension is $N = 4.2 \times 10^6$ , the node feature dimension is 256, the edge feature dimension is 12, the number of training iteration rounds is set to 300, and the minimum validation loss strategy is chosen to save the model snapshot in each round. The model training loss function is defined as weighted cross-entropy loss:

$$L = -\sum_{i=1}^{N} w_i \cdot \left[ y_i \log(\hat{y}_i) + (1 - y_i)\log(1 - \hat{y}_i) \right] \qquad (3)$$

Where $\hat{y}_i$ denotes the model output probability, $y_i$ is the real label, and $w_i$ is the weighting factor of user behavior, which improves the generalization ability under the unbalanced sample distribution. The inference stage adopts Graph Mini-Batch Inference (GMI) strategy, which is based on multi-channel parallel subgraph sampling and caching mechanism to control the upper limit of explicit memory, and the inference path adopts k-hop Neighborhood Expansion (NExpansion) strategy:

$$H^{(l+1)} = \sigma\left( \sum_{j \in N(i)} \frac{1}{\sqrt{d_i d_j}} W^{(l)} H_j^{(l)} \right) \qquad (3)$$

Where, $H^{(l)}$ is the node embedding of layer $l$ , $N(i)$ is the set of neighboring nodes of node $i$ , $W^{(l)}$ is the weight matrix of layer $l$ , and $d_i$ is the node degree. The inference process maintains a time window of Δt = 6 hours, a node sampling rate of 15%, a cache window depth of 2 hops, and a GPU graphics memory footprint controlled within 36.2 GB[7] . As shown in Fig. 4, the training and inference process is presented in a modularized way, which specifies the data input path, model loading strategy and multi-batch distributed computing process.

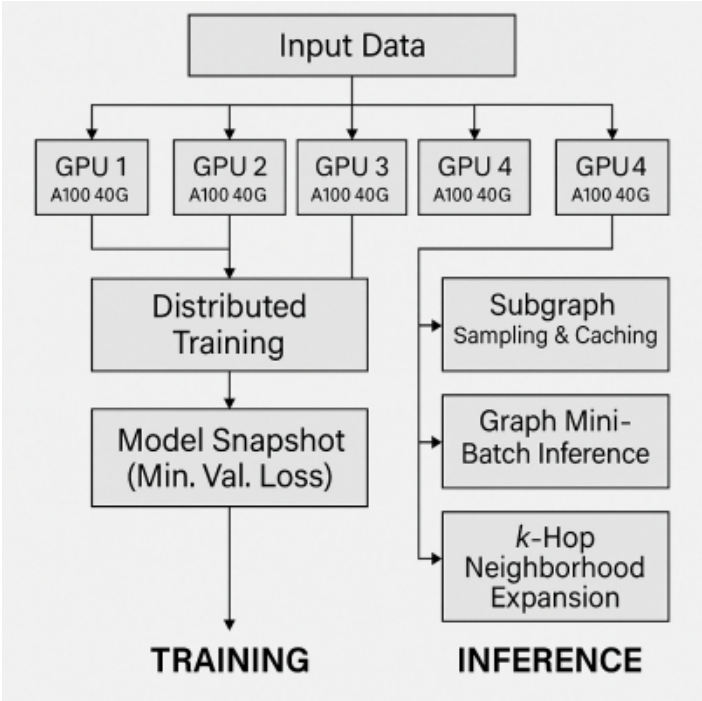

Figure 4. Flowchart of GNN model training and inference

## Experimental Results and Analysis

### Experimental Data Set

The experimental dataset is derived from ad placement logs from three mainstream platforms (Platform-A, Platform-B, and Platform-C), covering 2,870 records from January to December 2023. Data sources include social platforms, online video platforms, and information apps, captured via API. Fields include Ad_ID, Platform_ID, User_ID, Behavior Type (click, browse, stay), Timestamp, Device_Type, OS_Version, Content_Tag, and Contextual Characteristics (time period, geography, network environment). User behaviors are divided into 12 categories and 37 types of advertisement tags, and the data are de-weighted and normalized to provide a high-quality basis for graph neural network building and sequence recommendation[8] .

### Model Training and Evaluation

For the constructed cross-platform advertisement recommendation graph neural network model, the experimental phase is divided using 70% of training set, 15% of validation set and 15% of testing set[9] . The training process is performed using Adam optimizer with the initial learning rate set to 0.001, the number of training rounds is fixed to 300, and the cross-entropy loss is monitored on the validation set to implement the early stopping mechanism. The evaluation metrics are selected as five core metrics: AUC, Accuracy, F1 value, Precision, and Recall to comprehensively measure the discriminative ability and generalization performance of the model. These metrics were chosen because they provide a balanced view of model performance across different scenarios, particularly in a multi-platform context. AUC is used to evaluate the model's ability to rank ads correctly, regardless of the threshold, which is crucial in a multi-platform setting where platforms may have different label distributions. Accuracy is included to assess the overall correctness of the model, while Precision and Recall are vital for understanding the trade-off between false positives and false negatives, which is especially important in advertising, where targeting the right audience is critical. F1 value combines both Precision and Recall, offering a single metric that balances the two, making it particularly useful in contexts where both false positives and false negatives need to be minimized. The evaluation is carried out simultaneously with the independent dimension of each platform and the combined dimension of the whole platform to ensure the model's adaptability to heterogeneous data, as shown in Table 2.

TABLE II. SUMMARY OF MODEL TRAINING AND EVALUATION RESULTS

| Assessment dimensions | Accuracy (%) | Accuracy (%) | Recall rate (%) | F1 value (%) | AUC |
|---|---|---|---|---|---|
| Platform-A (test set) | 87.3 | 85.9 | 83.2 | 84.5 | 0.921 |
| Platform-B (test set) | 89.1 | 88.2 | 86.5 | 87.3 | 0.937 |
| Platform-C (test set) | 86.5 | 84.7 | 82.1 | 83.4 | 0.915 |
| Cross-platform merger (overall) | 88 | 86.6 | 84 | 85.3 | 0.931 |

According to the results in Table 2, the model demonstrates stable predictive performance on each platform test set, in which Platform-B has the highest AUC value of 0.937, indicating that it has stronger discriminative power in the recognition of advertising behaviors on this platform. Platform-A and Platform-C are slightly lower in terms of accuracy and F1 index. The analysis found that the distribution of advertisement label categories in these two platforms is more uneven, which leads to a decrease in the model's fitting ability on a few categories of samples.

To address this issue, we implemented and compared two mitigation strategies during training: (1) applying a weighted cross-entropy loss, where underrepresented ad labels were assigned higher weights (e.g., frequent label = 1.0, rare label = 1.8); and (2) oversampling minority classes to at least 1.5× their original count. Comparative experiments on Platform-C showed that weighted loss improved the F1 score from 83.4% to 84.6%, while oversampling enhanced Recall from 82.1% to 84.3%, albeit with a slight drop in Precision. These results suggest that label imbalance significantly affects performance, and a combination of weighting and sampling may offer more robust improvements.The combined dimension evaluation shows that the model maintains high consistency and robustness under the overall cross-platform data structure, and the F1 value reaches 85.3%, which indicates that the GNN architecture is well adapted to cope with the problems of platform heterogeneity and advertisement diversity[10] .

### Analysis of Experimental Results

The experimental evaluation results reflect that the model has high consistency and discriminative ability in multi-platform advertisement recommendation scenarios, but there are still some differences in the performance in terms of the diversity of user behaviors and the heterogeneity of platform contents. Combined with the statistics of five core indexes on each platform test set after training (see Table 3), it can be found that Platform-B outperforms Platform-A and Platform-C in AUC and F1 values. This suggests that the model demonstrates stronger generalization ability on platforms with clearer structural labels and more stable user behavioral patterns. As shown in Fig. 1, Platform-B's higher AUC and F1 values highlight its better ability to accurately predict user behaviors and recommend ads effectively, especially in environments with well-defined user actions. In contrast, Platform-C's performance fluctuates due to its higher label density and more complex ad types, leading to lower consistency in recommendation accuracy. The recall and precision of Platform-C fluctuates greatly, showing that the model has better generalization ability in the face of complex ad types and fragmented ads. the model's recognition stability decreases when facing complex ad types with fragmented behavioral sequences. In addition, although the overall merged results are high in all indicators, they imply the local error balance caused by the distribution differences between platforms, and further optimizing the model's ability to capture features of multi-platform differences will be the direction of subsequent improvement.

TABLE III. COMPARATIVE ANALYSIS OF EXPERIMENTAL INDICATORS OF EACH PLATFORM

| Assessment platforms | Accuracy (%) | Accuracy (%) | Recall rate (%) | F1 value (%) | AUC value | Average behavioral sequence length | Number of ad types | Label density (labels/advertiseme |
|---|---|---|---|---|---|---|---|---|
| | | | | | | | | nts) |
| Platform-A | 87.3 | 85.9 | 83.2 | 84.5 | 0.921 | 12.4 | 28 | 2.3 |
| Platform-B | 89.1 | 88.2 | 86.5 | 87.3 | 0.937 | 15.8 | 24 | 1.7 |
| Platform-C | 86.5 | 84.7 | 82.1 | 83.4 | 0.915 | 10.3 | 37 | 3.1 |
| Cross-platform merger | 88 | 86.6 | 84 | 85.3 | 0.931 | 13.2 | 37 | 2.5 |

From the data in Table 3, Platform-B performs optimally in all indicators, with an accuracy of 89.1%, an AUC value of 0.937, and an F1 value of 87.3%, thanks to its longest average behavioral sequence length (15.8), which enables the model to learn the user interest migration paths more adequately. Although Platform-C includes the most diverse ad types (37 categories) and the highest label density (3.1), it has the lowest F1 value and AUC, indicating that label redundancy and sample sparsity interfere with the model's discriminative performance. After merging the data, all the indicators are at a high level, indicating that the model has good stability under the overall multi-platform data fusion.

## Conclusion

The recommendation system for cross-platform ad campaigns effectively improves the accuracy of ad delivery through a graph neural network model, especially when dealing with heterogeneous data and diverse ad content, demonstrating strong adaptability. The experimental results show that the behavioral differences between different platforms and the density of ad labels have a significant impact on the performance of the model, especially in Platform-B, where the model exhibits the best accuracy and AUC value. Subsequent work should further optimize the multi-platform data fusion by exploring advanced techniques such as transfer learning to align features across platforms, semi-supervised learning to leverage unlabeled data for improving feature extraction, and domain adaptation methods to bridge the gap between different platform behaviors. Additionally, improving the model's ability to recognize complex ad types and fragmented user behaviors can be achieved through incorporating multi-modal data, such as text, images, and behavioral sequences, to better capture diverse content features and user interactions. These approaches will contribute to more accurate and efficient cross-platform ad recommendations.

In this article, Xiang Li and Xinyu Wang are the co-first authors.

## Reference

[1] Heo H, Lee S. Consumer Information-Seeking and Cross-Media Campaigns: An Interactive Marketing Perspective on Multi-Platform Strategies and Attitudes Toward Innovative Products [J]. Journal of Theoretical and Applied Electronic Commerce Research, 2025, 20(2): 68.

[2] Rizgar F, Zeebaree S R M. The Rise of Influence Marketing in E-Commerce: a Review of Effectiveness and Best Practices [J]. East Journal of Applied Science, 2025, 1(1): 18-34.

[3] Zhang K, Xing S, Chen Y. Research on Cross-Platform Digital Advertising User Behavior Analysis Framework Based on Federated Learning[J]. Artificial Intelligence and Machine Learning Review, 2024, 5(3): 41-54.

[4] Akmal T, Bilal M Z, Raza A. DVC Digital Video Commercial of Pakistani Food Advertising Industry: A Qualitative Analysis of Contemporary Digital Trends in Advertising Industry [J]. Contemporary Journal of Social Science Review, 2025, 3(1): 808-822.

[5] Kasih E W, Benardi B, Ruslaini R. The power of Sequence: A Qualitative Analysis of Consumer Targeting and Spillover Effects in Social Media Advertising [J]. International Journal of Business, Marketing, Economics & Leadership (IJBMEL), 2024, 1(4): 28-42.

[6] Hatzithomas L, Theodorakioglou F, Margariti K, et al. Cross-Media Advertising Strategies and Brand Attitude: The Role of Cognitive Load [J]. International Journal of Advertising, 2024, 43(4): 603-636.

[7] Carah N, Hayden L, Brown M G, et al. Observing "tuned" advertising on digital platforms [J]. Internet Policy Review, 2024, 13(2): 1-26.

[8] Hassan H E. Social Media Advertising Features that Enhance Consumers' Positive Responses to Ads [J]. Journal of Promotion Management, 2024, 30(5): 874-900.

[9] Chen X J, Chen Y, Xiao P, et al. Mobile ad fraud: Empirical patterns in publisher and advertising campaign data [J]. International Journal of Research in Marketing, 2024, 41(2): 265-281.

[10] Kwon H B, Lee J, Brennan I. Complex interplay of R&D, advertising and exports in USA manufacturing firms: differential effects of capabilities [J]. Benchmarking: An International Journal, 2025, 32(2): 459-491.